\documentclass[journal,twoside,web]{ieeecolor}
\usepackage{bm}
\usepackage{tmi}
\usepackage{cite}
\usepackage{amsmath,amssymb,amsfonts}
\usepackage{algorithmic}
\usepackage{graphicx}
\usepackage{textcomp}
\usepackage{amssymb,mathrsfs,amsmath}
\usepackage{graphicx}
\usepackage{booktabs}
\usepackage{multirow}
\usepackage{multicol}
\usepackage{array}

\usepackage[colorlinks,linkcolor=blue]{hyperref}

\def\BibTeX{{\rm B\kern-.05em{\sc i\kern-.025em b}\kern-.08em
    T\kern-.1667em\lower.7ex\hbox{E}\kern-.125emX}}
\begin{document}
\title{StyleSeg~V2: Towards Robust One-shot Segmentation of Brain Tissue via Optimization-free Registration Error Perception
}
\author{Zhiwei Wang, Xiaoyu Zeng, Chongwei Wu, Jinxin Lv, Xu Zhang, Wei Fang and Qiang Li, \IEEEmembership{Member, IEEE}
\thanks{Zhiwei Wang, Xiaoyu Zeng, Chongwei Wu, Jinxin Lv and Qiang Li are with Britton Chance Center for Biomedical Photonics, Wuhan National Laboratory for Optoelectronics, Huazhong University of Science and Technology, Wuhan, Hubei 430074, China. Xu Zhang and Wei Fang are with Wuhan United Imaging Healthcare Surgical Technology Co., Ltd., Wuhan, 430074, China. Zhiwei Wang and Xiaoyu Zeng are the co-first authors contributing equally to this work. Qiang Li is the corresponding author. (email:liqiang8@hust.edu.cn)}}
\maketitle

\begin{abstract}
One-shot segmentation of brain tissue requires training registration-segmentation (reg-seg) dual-model iteratively, where reg-model aims to provide pseudo masks of unlabeled images for seg-model by warping a carefully-labeled atlas. However, the imperfect reg-model induces image-mask misalignment, poisoning the seg-model subsequently. Recent StyleSeg bypasses this bottleneck by replacing the unlabeled images with their warped copies of atlas, but needs to borrow the diverse image patterns via style transformation. Here, we present StyleSeg~V2, inherited from StyleSeg but granted the ability of perceiving the registration errors. The motivation is that \emph{good registration behaves in a mirrored fashion for mirrored images}. Therefore, almost at no cost, StyleSeg~V2 can have reg-model itself ``speak out'' incorrectly-aligned regions by simply mirroring (symmetrically flipping the brain) its input, and the registration errors are symmetric inconsistencies between the outputs of original and mirrored inputs. Consequently, StyleSeg~V2 allows the seg-model to make use of correctly-aligned regions of unlabeled images and also enhances the fidelity of style-transformed warped atlas image by weighting the local transformation strength according to registration errors. The experimental results on three public datasets, i.e., OASIS, CANDIShare, and MM-WHS 2017 demonstrate that our proposed StyleSeg~V2 outperforms other state-of-the-arts by considerable margins, and exceeds StyleSeg by increasing the average Dice by 2.0\%, 2.4\%, 1.7\% on the three datasets, respectively.
\end{abstract}


\section{Introduction}
\label{sec:introduction}
\IEEEPARstart{B}{rain} tissue segmentation is a fundamental technique in clinical practice, which enables accurate disease diagnosis and treatment planning \cite{geuze2005mr}. The existing fully supervised segmentation methods can achieve promising segmentation accuracy with sufficient well-annotated data. However, the sophisticated brain structure makes the 3D manual annotation of brain MRI images labor-intensive and error-prone. Therefore, training a segmentation method using only a few labeled images is of great practical significance.

\begin{figure}[t]
\centerline{\includegraphics[width=0.95\columnwidth]{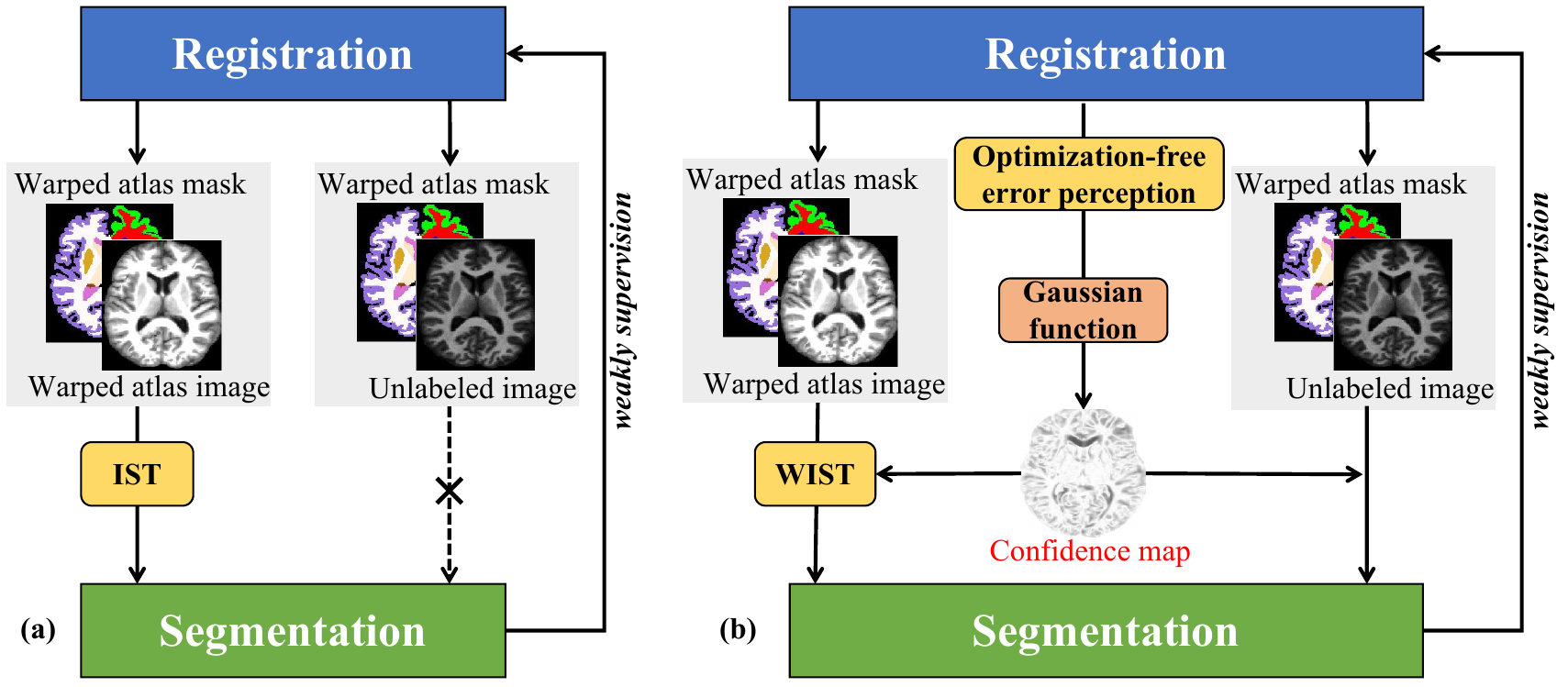}}
    \caption{The schematic diagram of (a) StyleSeg and (b) StyleSeg~V2.}
    \vspace{-0.4cm}
	\label{introduction}
\end{figure}

A straightforward solution is called atlas-based segmentation \cite{lorenzo2002atlas,lotjonen2010fast}. The key idea is to first establish a pixel-wise correspondence between a carefully-labeled atlas image (moving image) and an unlabeled image (fixed image) through unsupervised registration based on image-level similarity, and then the mask of atlas can be propagated as the segmentation result. 

However, the registration model pays more attention to a overall spatial accordance instead of the specific brain tissue of interest, which therefore leads to sub-optimal segmentation results ~\cite{he2020deep}. In comparison, a more effective paradigm proposed recently is to jointly learn two models for weakly supervised registration and semi-supervised segmentation, which forms a dual-model iterative learning~\cite{he2020deep,beljaards2020cross,zhao2019data,he2022learning}. Concretely, a registration model (reg-model) is first trained purely relying on target-agnostic image-level similarity, providing pseudo masks of unlabeled images for supervising a segmentation model (seg-model); the learned seg-model then predicts refined masks to weakly supervise the reg-model, forcing it to pay more attention to the tissue of segmentation interest; in the next iteration the reg-model can warp better and thus enhance the seg-model predictably. 

Despite their success, a common bottleneck occurs when using the reg-model to generate pseudo masks of unlabeled images. That is, the reg-model is certainly imperfect, thus yielding incorrectly-aligned regions in each image-mask pair. The seg-model could partly tolerate these misaligned regions thanks to the inductive ability, but eventually be unbearable as they iteratively accumulate.

To address the above bottleneck, StyleSeg \cite{lv2023robust} was proposed as shown in Fig.~\ref{introduction}(a). It rejects the image-mask pairs containing registration error, and takes a bypass by using the warped copies of atlas-mask pairs, that is, replacing the unlabeled image with the warped atlas image. However, StyleSeg has a major issue: the seg-model always ``sees'' the copies of the same atlas, ignoring the valuable information carried by the unlabeled images. Although StyleSeg designs image-aligned style transformation (IST) to borrow the varied image patterns from the unlabeled images, the registration error could also degrade the fidelity of the style-transformed atlas copies. Therefore, bypassing the registration error is just a compromise and could cause other implicit troubles eventually. A few studies, e.g., DeepRS \cite{he2020deep}, have tried to predict the registration error by mostly resorting to complex extra models like GAN, which, however, intensifies training costs on top of learning dual-model on 3D paired images.

In this paper, we introduce StyleSeg~V2 as the next generation of StyleSeg. This upgraded version not only enhances the reliability of style-transformed atlas images but also permits the unlabeled images to join the training process effectively, as illustrated in Fig.~\ref{introduction}(b). Specifically, StyleSeg~V2 ``awakens'' the inherent yet often overlooked capability of the registration model to perceive registration error without relying on any learnable parameters. Motivated by the observation that the two symmetrical brain hemispheres often exhibit asymmetric registration error, StyleSeg~V2 generates mirrored images by swapping the left and right brain parts symmetrically, and identifies error on the original image by checking the symmetrical consistencies of predictions on the mirrored one. By use of registration error, StyleSeg~V2 reinvents IST as weighted IST (WIST) by reducing the style transformation force on the atlas image for those misaligned local regions, and moreover selects the low-error regions from the unlabeled images to directly constrain seg-model.

To summarize, our contributions are listed as follows:
\begin{itemize}
\item[$\bullet$] We propose StyleSeg~V2 for robust one-shot brain segmentation, which distinguishes itself from the original StyleSeg by a power of perceiving registration errors almost effortlessly.
\item[$\bullet$] We propose two enabling techniques on top of the perceived registration errors, i.e., weighted image-aligned style transformation and confidence guided Dice loss. The former enables StyleSeg~V2 to ``see'' more diverse and high-fidelity style-transformed atlas image in the bypass training, and the latter boosts StyleSeg~V2 directly using the correctly-aligned regions of unlabeled images.
\item[$\bullet$] The experimental results on three public datasets demonstrate the superiority of StyleSeg~V2 in both brain segmentation and registration. Compared to the original version, StyleSeg V2 improves the segmentation performance by increasing the average Dice by 2.0\%, 2.4\%, and 1.7\% on the three datasets, respectively.
\end{itemize}

\section{RELATED WORK}

\begin{figure*}[!t]
\centerline{\includegraphics[width=0.95\textwidth]{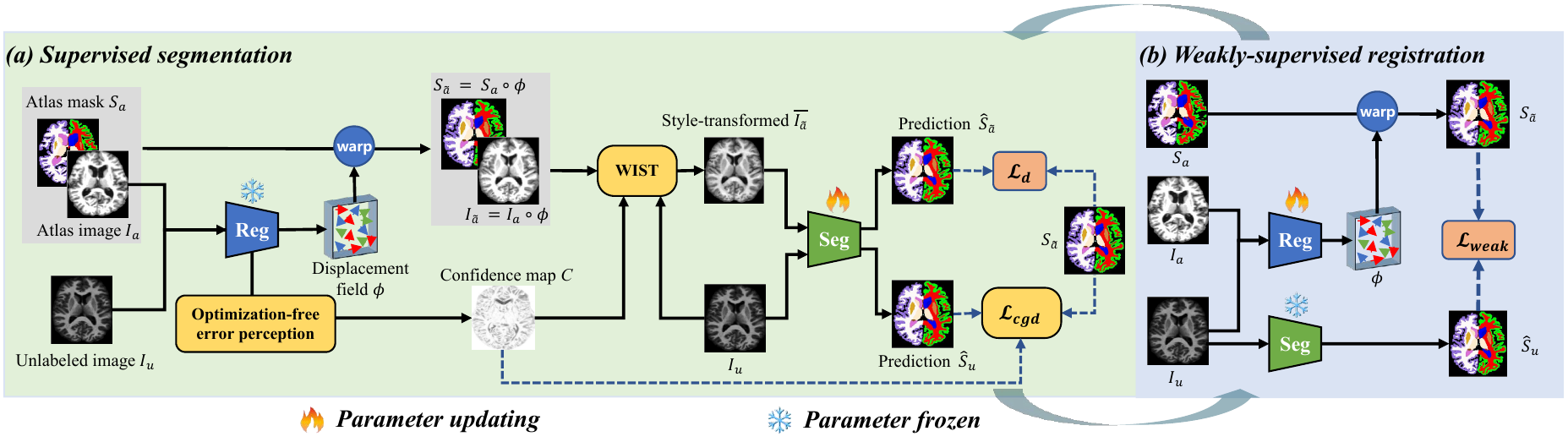}}
\caption{The overview of iterative training in StyleSeg~V2. (a) We first frozen reg-model to predict a pseudo mask (warped atlas mask) and a registration confidence map for supervising seg-model on both warped atlas image and unlabeled one. Weighted image-aligned style transformation style-transformed (WIST) increases both diversity and fidelity of the warped atlas image, and confidence guided Dice loss $\mathcal{L}_{cgd}$ constrains the seg-model using the aligned regions of the unlabeled image. (b) We then fix the seg-model to produce a refined prediction of unlabeled image, which can be further exploited to weakly supervise the reg-model, triggering a new round of iteration}
	\label{framework}
\end{figure*}

\subsection{Atlas-based segmentation}
To alleviate the negative impact caused by the lack of annotations, a simple approach is to use the atlas-based segmentation method. The standard atlas-based approach treats segmentation as a registration problem by mapping the few atlas images and their labels onto the unlabeled images. Some traditional registration methods are often employed. For instance, Collins \textit{et al.} \cite{collins1995automatic} extracted the image features based on blurred image intensity and image gradient magnitude to perform registering atlas to unlabeled images. J.W. Suh \textit{et al.} \cite{suh2013automatic} utilized two different characterized registration methods ANTS \cite{avants2008symmetric} and Elastix \cite{klein2009elastix} to deform each atlas in multi-atlas segmentation and combined the warped segmentations into a single consensus using a label fusion strategy. However, the conventional registration methods have high computing cost caused by the iterative optimization, and are hard to accurately align brain images with complex anatomical structures.

Recently, many learning-based registration methods have achieved promising performance, which also promote the capability of atlas-based segmentation. Wang \textit{et al.} \cite{wang2020lt} introduced cycle-consistency-based supervision to constrain the registration network to achieve consistent results after a forward and a backward warping. Ding \textit{et al.} \cite{ding2022aladdin} proposed an atlas building method and a pairwise alignment strategy which evaluated the similarity in both moving image (atlas) space and fixed one. Some works \cite{dual-MICCAI,mok2020large,zhao2019unsupervised,Zhao_2019_ICCV,hu2022recursive,lv2022joint} aim to enhance the registration reliability by decomposing the one-step warping into multiple steps in a progressive and/or coarse-to-fine manner. Despite the robustness and great generality in different scenarios, the atlas-based segmentation methods are prone to an unsatisfying result since they mostly care about the global image-level registration, while the  small and subtle anatomical structures may be rarely focused.

\subsection{Dual-model based one-shot brain segmentation}
One-shot brain segmentation aims at using only a single labeled image (atlas) to train a segmentation model for brain tissues. The dual-model iteration strategy is one of the most typical and effective solutions. For example, DeepAtlas~\cite{xu2019deepatlas} treated the reg-model as a data augmentation scheme to provide image-mask pairs for segmentation, and then utilized the seg-model to conversely guide the reg-model learning. Despite their success, the main issue is that the reg-model inevitably causes misalignment between the unlabeled image and registration-provided pseudo mask, misguiding the seg-model subsequently. To address this, some methods was proposed to apply a style transformation on atlas to equip it with more diverse patterns and directly used the warped copies of atlas-mask pair to train the seg-model. For instance, Zhao \textit{et al.} \cite{zhao2019data} appended an extra appearance transform model to learn the style discrepancy from unlabeled images to atlas ones and then added the sampled style to original atlas images. As they introduced additional learning tasks, the whole training process was more cumbersome and computative. He \textit{et al.} \cite{he2022learning} followed such strategy by utilizing the backward registration from randomly sampled unlabeled image to atlas image, and then they regarded the subtraction image as the style patterns. However, simply adding the sampled subtraction image could damage the subtle structures of brain tissues and the image-label consistency. In order to make the process of style transformation more accurate and computation efficient, our previous work StyleSeg~\cite{lv2023robust} considered the Fourier amplitude spectrum of image mainly containing statistical information of style, and linearly mixed up the amplitude components between unlabeled image and warped one to generate the style-transformed atlas image. However, the style transformation is still negatively affected by registration error more or less, which makes the generated warped image contains visual artifacts due to imperfect spatial alignment. Besides, since the StyleSeg only utilized the warped atlas image in segmentation learning, the seg-model wastes the valuable information carried by the unlabeled images, and thus could overfit to the image patterns of the atlas.

\subsection{Registration error perception}Registration error perception aims to find where the misalignment occurs in the results of a registration model. Some methods \cite{eppenhof2017supervised,eppenhof2018error,sokooti2021hierarchical,bierbrier2023towards} tried to train a supervised model using synthetic data with known registration errors. However, the scalability of such methods are limited as the deformation in medical scenes is complex and hard to simulate. In view of this, there were a few methods \cite{christensen2001consistent,meister2018unflow,mok2022unsupervised}, as known as forward-backward consistency, proposed a cycle-form registration from the moving image to the fixed, and then back to the moving itself. They assumed that a voxel with good registration will come back to the original position precisely in the cycle registration. However, such manner only estimates the total error of twice registration (moving-fixed and fixed-moving), while the error in each of the two paths is still unknown. To perceive the error of once registration, Saygili \textit{et al.} \cite{saygili2015confidence} constructed a voxel-level cost space between the registered image pairs and designed an objective function to calculate the confidence map via optimization. DeepRS \cite{he2020deep} considered the reg-model as generator, and trained a discriminator to produce a pixel-wise confidence map based on adversarial learning. Nevertheless, these methods introduce an additional optimization task, resulting in heavy computing overhead and making the dual-model learning process more cumbersome.

To mitigate the cost of error perception, we in this paper seek an optimization-free method to perceive the single direction registration error. Inspired by the observation that the symmetric brain structures often exhibit asymmetric registration performance, we propose to estimate the registration error by comparing the differences between the deformation fields in the original and mirror data spaces. Such process requires no optimization overhead.

\section{Methods}
\subsection{Overview of iterative learning in StyleSeg~V2}
Fig. \ref{framework} illustrates our improved iterative and complementary framework. We split the whole framework into registration and segmentation parts. In registration module, we utilize the mirror error to  approximate the registration error and acquire the confidence map. Utilizing the confidence map, we can generate style-diverse image samples with less artifacts via WIST and also combine the real unlabeled image information via our proposed loss function $\mathcal{L}_{cgd}$ in segmentation learning. In the following, we first introduce the procedure of registration error estimation via mirror error (Sec. \ref{mirror_error}), then detail both WIST and $\mathcal{L}_{cgd}$ (Sec. \ref{l_{d}}\&Sec. \ref{Lcgd}), and at last give implementation and training details (Sec. \ref{setting}).

\subsection{Optimization-free registration error perception}\label{mirror_error}
\begin{figure}
\centerline{\includegraphics[width=\linewidth]{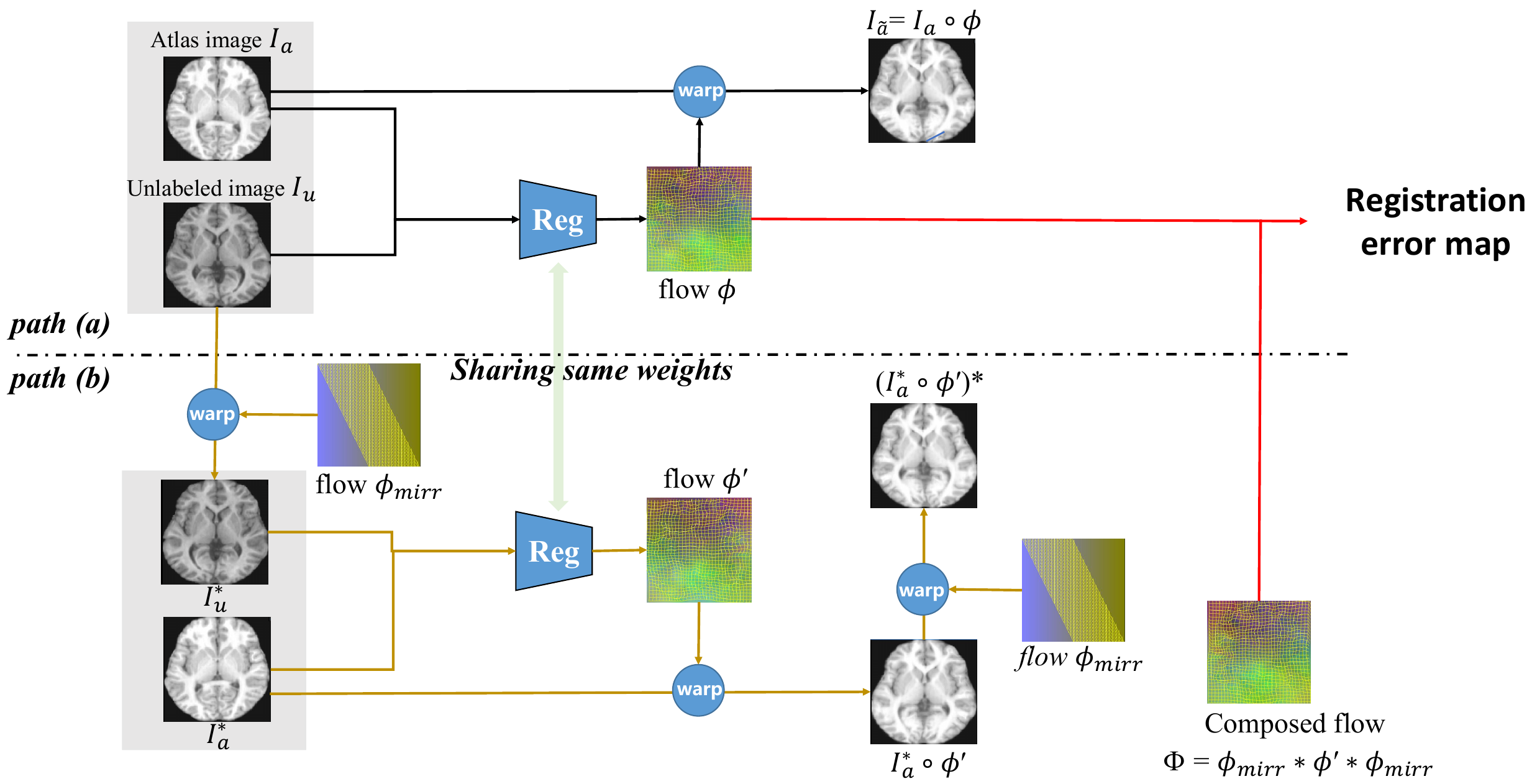}}
\caption{The calculation of the registration error map formulated in Eq.~\ref{eq:2}. Registration path (a): direct registraion, to produce deformation flow $\phi$; Registration path (b): flipping-registration-flipping, to produce composed flow $\Phi$. The registration error is the L2 distance between $\phi$ and $\Phi$.}
	\label{flip}
\end{figure}

Given the 3D atlas image $I_{a}$ and an unlabeled one $I_{u}$, the reg-model predicts a displacement field $\phi$ and has $I_{\tilde{a}} = I_{a} \circ \phi \rightarrow I_{u}$, where $\circ$ is a simplified notation of warping, $\rightarrow$ points to the warping target, and $I_{\tilde{a}}$ is the warped image. We find that for symmetrical brain structures, the poor registration results exhibit strong non-symmetric characteristics at the left and right sides. Based on this phenomenon, StyleSeg~V2 can force a learned reg-model to identify its own prediction errors without other extra optimization. Fig. \ref{flip} shows the overall procedure.

Specifically, StyleSeg~V2 symmetrically flips the image along the sagittal axis in the 3D space to generate a fake image where the left and right brain parts are swapped. We denote such operation as ``mirror'' operation and distinguish the mirrored image using a superscript $*$. The atlas and unlabeled images are mirrored to get ($I^{*}_{a}$, $I^{*}_{u}$) and fed into the reg-model to predict a new displacement field $\phi^{'}$, yielding a warped mirrored atlas image $I^{*}_{a} \circ \phi^{'} \rightarrow I^{*}_{u}$. We next mirror the resulting image back and combing the two times of warping as:
\begin{equation}
(I^{*}_{a} \circ \phi^{'})^{*} \rightarrow (I^{*}_{u})^{*} = I_{u} \leftarrow  I_{a} \circ \phi
\label{eq:1}
\end{equation}

Therefore, we have $(I_{a}^{*} \circ \phi')^{*} \rightleftarrows I_{a} \circ \phi$. A straightforward way of quantifying the registration errors is to calculate the voxel-level difference between $(I^{*}_{a} \circ \phi^{'})^{*}$ and $I_{a} \circ \phi$. However, such manner could be too intensity sensitive to fully reveal the accurate spatial misalignment. To avoid this, StyleSeg~V2 calculates the registration errors purely based on the displacement fields instead of voxel intensities. 

The ``mirror'' operation can be mathematically expressed as a fixed displacement field $\phi_{mirr}$ defined as $\phi_{mirr}(x,y,z) = (N_{x}-2x, 0, 0)$, where $x$ is on the axis perpendicular to the symmetric plane, i.e., the sagittal plane, and $N_{x}$ is the axis length. Therefore, Eq.~(\ref{eq:1}) can be rewritten as $((I_{a} \circ \phi_{mirr}) \circ \phi^{'}) \circ \phi_{mirr} \leftrightarrows I_{a} \circ \phi$. According to \cite{zhao2019unsupervised}, the multiple warping can be simplified as a single warping, and the relationship is that $((I_{a} \circ \phi_{mirr}) \circ \phi^{'}) \circ \phi_{mirr} = I_{a} \circ(\phi_{mirr}+(\phi^{'}+\phi_{mirr} \circ \phi^{'}) \circ \phi_{mirr})$. Thus, we have the final equation:
\begin{equation}
I_{a} \circ \Phi \leftrightarrows I_{a} \circ \phi
\label{eq:3}
\end{equation}
where $\Phi = \phi_{mirr} +(\phi^{'} + \phi_{mirr} \circ \phi^{'}) \circ \phi_{mirr}$ is a displacement field equivalent to warping by use of $\phi_{mirr}$, $\phi^{'}$ and $\phi_{mirr}$ sequentially.

Using Eq.~(\ref{eq:3}), the error map $E$ of the original registration from the atlas to unlabeled image can be easily obtained by calculating the distance between the displacement vectors of $\Phi$ and $\phi$, which is formulated as follows:
\begin{equation}
E = \| \Phi - \phi \|_{2}
\label{eq:2}
\end{equation}

To normalize the registration error, we further convert $E$ into a confidence map $C$ ranged from 0 to 1 by using a Gaussian transfer function on the error map $E$, that is:
\begin{equation}\label{errpr definition}
    C =exp(-\frac{E^2}{2\sigma^2}).
\end{equation} 
where $\sigma$ is the standard deviation of $E$. Each entry of $C$ indicates a probability of the two images being well-aligned with each other at this position.

\subsection{Weighted image-aligned style transformation for training with atlas}
\begin{figure*}[!t]
\centerline{\includegraphics[width=0.95\textwidth]{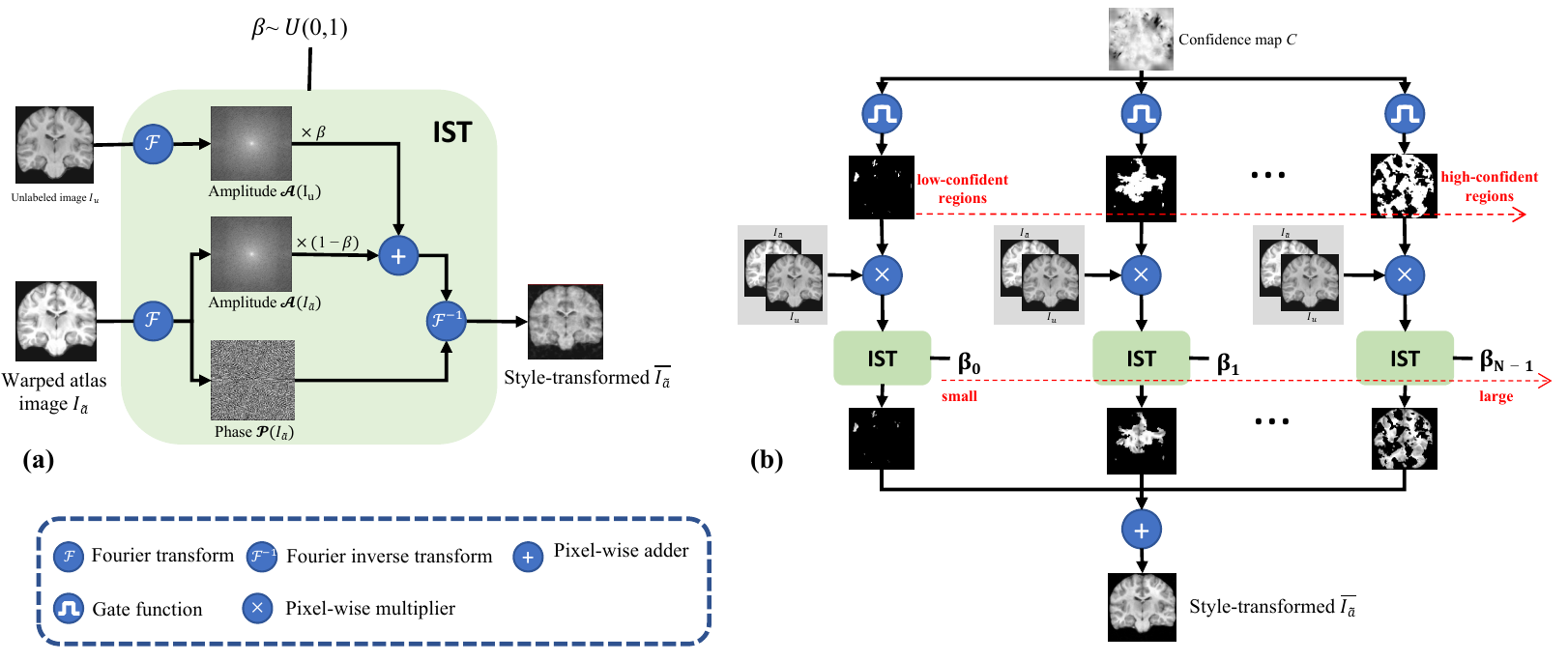}}
    \caption{(a) The detail of original IST, which swaps the Fourier amplitude component of the warped atlas with that of the unlabeled images to transfer the style. (b)The detail of weighted IST (WIST), which divides the warped atlas $I_{\tilde{a}}$ and unlabeled $I_{u}$ images into multiple sub-images with different confidence intervals, and applies lesser style-transformation strength on the sub-images with lower confidence, and vice versa. The style-transformed sub-images are added together to give the final WIST-transformed warped atlas image $\overline {I_{\tilde{a}}}$.}\label{WIST}
\end{figure*}

\label{l_{d}}
To avoid lacks of diversity, the original StyleSeg utilized image-aligned style transformation (IST) between $I_{\tilde{a}}$ and $I_{u}$ to produce style-diverse image $\overline {I_{\tilde{a}}}$ for supervising the seg-model. As shown in Fig. ~\ref{WIST}(a),  the principle of IST can be formulated as follows:
\begin{equation}\label{Fourier}
\resizebox{0.9\linewidth}{!}{
    $\overline {I_{\tilde{a}}} = \mathcal{F}^{ -1} \left( \left( \beta \times \mathcal{A}({I_{u}}) + (1-\beta) \times \mathcal{A}({I_{\tilde{a}}}) \right)e^{-j \times \mathcal{P}\left( I_{\tilde{a}}\right)} \right).$
}
\end{equation}
where $\mathcal{F}^{ -1}$ is inverse Fourier transformation, $\mathcal{A}$ and $\mathcal{P}$ represent the amplitude and phase spectrum, and $\beta$ controls the strength of style transformation. 

Utilizing IST, the generated $\overline {I_{\tilde{a}}}$ contains the style patterns of unlabeled image and also has correct spatial correspondence with the pseudo mask $S_{\tilde{a}} = S_{a} \circ \phi$. Therefore, the seg-model can be trained with the image-mask pairs ($\overline {I_{\tilde{a}}}$,$S_{\tilde{a}}$).

\begin{figure}[t]
\centerline{\includegraphics[width=\linewidth]{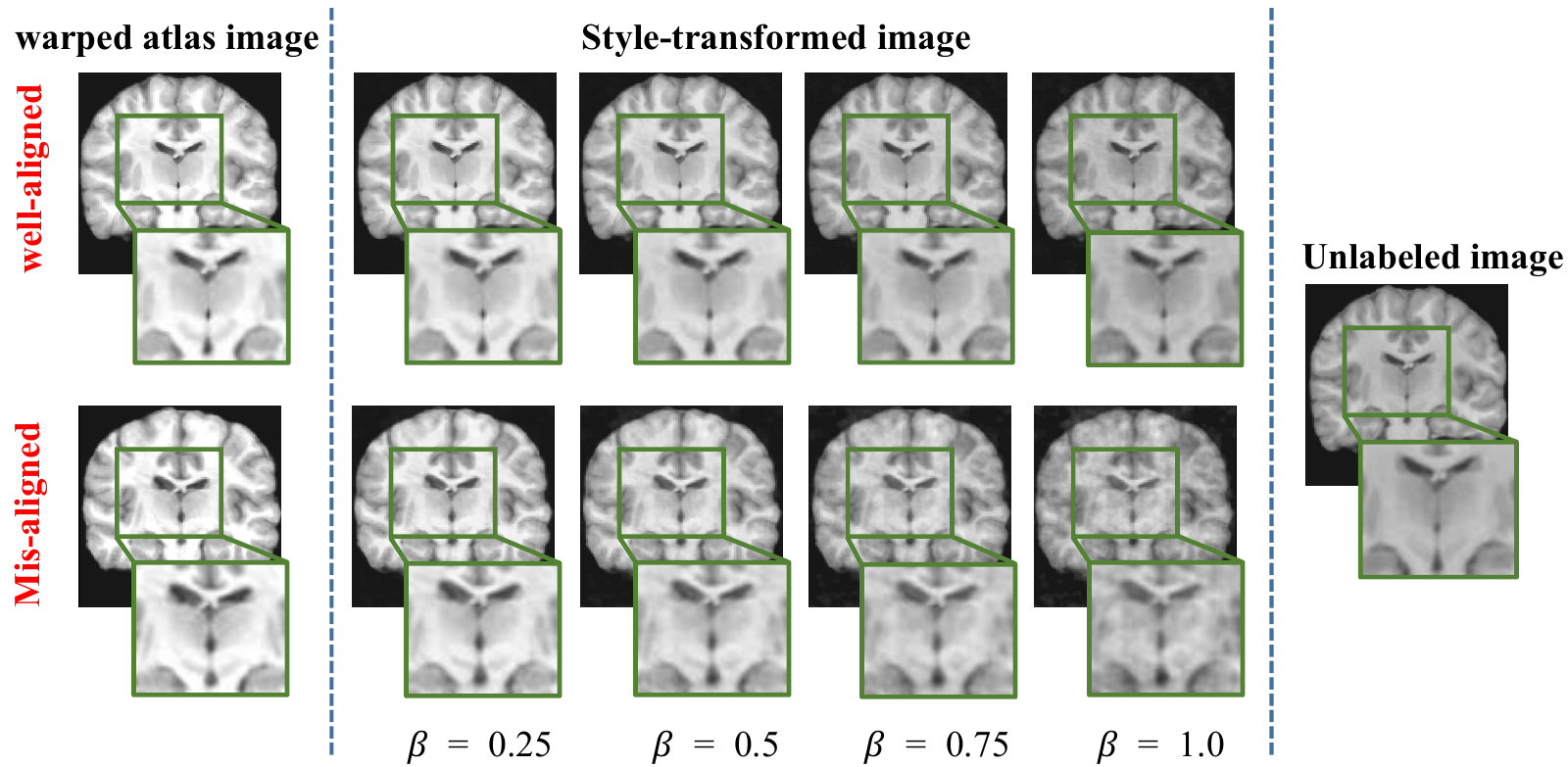}}
    \caption{The flaw of IST: artifacts appear if the images are mis-aligned due to imperfect registration, and become severer as the transformation strength goes higher.}\label{IST}
\end{figure}

However, mixing $\mathcal{A}(I_{\tilde{a}})$ and $\mathcal{A}(I_{u})$ induces visual artifacts especially when $I_{\tilde{a}}$ and $I_{u}$ are incorrectly aligned. As evidenced in Fig.~\ref{IST}, it causes visual artifacts for misaligned images, and this side effect gets worse on those misaligned regions as the transformation strength $\beta$ goes higher.

To address this, StyleSeg~V2 adopts the most intuitive but effective solution that dynamically lowers the style transformation strength $\beta$ according to the registration errors, i.e., the confidence map $C$. Therefore, an improved IST is proposed and called weighted image-aligned style transformation (WIST) as depicted in
Fig.~\ref{WIST}(b). Specifically, we firstly acquire $N$ binary masks utilizing a series of gate functions through the formula: 
\begin{equation}\label{WIST caculation}
\begin{aligned}
    M_{n}(x,y,z)= 
    \begin{cases}
    1, \quad \text{if} \enspace \frac{n}{N} \leq C(x,y,z) < \frac{n+1}{N} \\
    0, \quad \text{otherwise}.
    \end{cases}
\end{aligned}   
\end{equation}
where $n=0,...,N-1$ and $(x,y,z)$ is a voxel index. 

Then we multiply the $I_{u}$ and $I_{\tilde{a}}$ with $M_{n}$ to separate them into multiple regions with different registration confidence. We perform IST on the $M_{n}$-extracted regions  using $\beta_{n}$, which is randomly sampled from the interval $[\frac{n}{N}, \frac{n+1}{N})$ and finally add all results together to obtain the weighted style-transformed atlas image $\overline {I_{\tilde{a}}}$.

Utilizing WIST, the generated $\overline {I_{\tilde{a}}}$ will have less artifacts even when the reg-model has large registration errors in the initial stage of iteration. Thus, we can trustingly feed $\overline {I_{\tilde{a}}}$ into seg-model to predict a mask $\hat{S}_{\tilde{a}}$ and constrain the seg-model with the loss $\mathcal{L}_{d} = -Dice(\hat{S}_{\tilde{a}},S_{\tilde{a}})$, where $Dice$(·, ·) is the Dice coefficient and can be formulated as:
\begin{equation}\label{Dice}
	Dice(A,B)=\frac{2|A\cap B|}{|A|+|B|}.
\end{equation}

\subsection{Confidence guided Dice loss of unlabeled images}\label{Lcgd}
The original StyleSeg did not directly use the unlabeled images for segmentation training since the pseudo mask $S_{\tilde{a}}$ and the unlabeled image $I_{u}$ spatially mismatch. Under the guidance of the registration confidence $C$, StyleSeg~V2 can suppress the mis-aligned regions in the segmentation results of $I_{u}$ and $S_{\tilde{a}}$, , and thus encourages the seg-model to more focus on the regions where the pseudo mask is more reliable. Therefore, the confidence guided Dice-loss is defined as follows for supervising the seg-model using unlabeled images:
 \begin{equation}\label{Lcgd calculation}
    \mathcal{L}_{cgd} = -Dice(C\hat{S}_{u}, CS_{\tilde{a}}).
\end{equation}
where $\hat{S}_{u}$ is the segmentation result of $I_{u}$ predicted by seg-model. Using the $\mathcal{L}_{cgd}$, the unlabeled images can also be utilized, which reduces the data waste and further boosts the seg-model.

\subsection{Implementation and training details}\label{setting}
\subsubsection{Registration constraints}
In the initial phase, StyleSeg~V2 trains the reg-model in an unsupervised manner using two losses $\mathcal{L}_{IC}$ and $\mathcal{L}_{smo}$. $\mathcal{L}_{IC}$ is a similarity measurement to constrain the consistency between the warped atlas image and fixed image. We use Normalization Local Correlation Coefficient (NLCC) to calculate the similarity, which is based on Person's Correlation Coefficient:
\begin{equation}\label{NLCC}
\begin{aligned}
    \rho\left(X, Y\right) =
    \frac{\sum(X-\bar{X})(Y-\bar{Y})}
    {\sqrt{\sum(X-\bar X)^{2}}\sqrt{\sum(Y-\bar Y)^{2}}}.
\end{aligned}
\end{equation}
where $X$ and $Y$ are two variables, e.g., images, and $\bar X$ and $\bar Y$ are mean of $X$ and $Y$ respectively.
  
Instead of directly calculating $\rho(I_{\tilde{a}}, I_{u})$, we first extract multiply patches from $I_{\tilde{a}}$ and $I_{u}$ separately using a $9\times9\times9$ sliding window with stride set to 1, and then average $-\left(\rho(X, Y)\right)^{2}$ for those patches as $\mathcal{L}_{IC}$:
\begin{equation}\label{NLCC_loss}
\begin{aligned}
    \mathcal{L}_{IC} = -\frac{1}{N}
    \sum\rho(I_{\tilde{a}}, I_{u}, i)^{2}
\end{aligned}
\end{equation}
where $\rho\left(X, Y, i\right)$ is the $\rho$ coefficient value for the i-th patches of $X$ and $Y$, and $N$ is the total number of voxels.

$\mathcal{L}_{smo}$ is a regular term to constrain the smoothness of displacement field, which is formulated as $\mathcal{L}_{Smo}=\left\| {\nabla\phi} \right\|_{2}^{2}$.

In the subsequent training of the reg-model, we additionally utilize a loss $\mathcal{L}_{weak}$ as an auxiliary weak-supervision term to improve the registration performance on the regions of interest. $L_{weak}$ constrains the pseudo mask $S_{\tilde{a}}$ of unlabeled image predicted by reg-model to be consistent with the refined one $\hat{S}_{u}$ predicted by the frozen seg-mode, that is, $\mathcal{L}_{weak} = -Dice(S_{\tilde{a}}, \hat{S}_{u})$. 

\begin{table*}
\centering
\renewcommand\arraystretch{1.1}
\caption{Comparison results of seg-model on the three datasets. The best performance is marked in bold}
\scalebox{1.0}{
\begin{tabular*}{0.90\textwidth}{lcccccc}
\toprule
\multirow{2}*{\centering Method} & \multicolumn{2}{c}{\centering OASIS} & \multicolumn{2}{c}{\centering CANDIShare} & \multicolumn{2}{c}{MM-WHS 2017}\\
\cmidrule(lr){2-3} \cmidrule(lr){4-5}\cmidrule(lr){6-7}
& Dice $\uparrow$ & HD $\downarrow$ & Dice $\uparrow$  & HD $\downarrow$ & Dice $\uparrow$  & HD $\downarrow$\\
\midrule
Brainstorm &$0.813\pm0.018$ &$1.651\pm0.327$ &$0.827\pm0.015$ &$2.133\pm0.415$ &$0.825\pm0.044$ &$7.455\pm2.354$\\
DeepAtlas &$0.819\pm0.018$ &$1.659\pm0.330$ &$0.828\pm0.014$ &$2.059\pm0.415$ &$0.849\pm0.039$ &$6.187\pm2.060$\\
PC-Reg-RT &$0.764\pm0.025$ &$2.846\pm0.533$ &$0.829\pm0.013$ &$2.260\pm0.278$ &$0.868\pm0.029$ &$5.364\pm1.917$\\
BRBS &$0.835\pm0.009$ &$1.622\pm0.270$ &$0.851\pm0.007$ &$1.882\pm0.305$ &$0.892\pm0.020$ &$5.551\pm2.084$\\
StyleSeg &$0.851\pm0.017$ &$\bm{1.509\pm0.307}$ &$0.839\pm0.012$ &$1.982\pm0.275$ &$0.886\pm0.024$ &$4.923\pm1.271$\\
StyleSeg~V2 &$\bm{0.868\pm0.004}$ &$1.513\pm0.247$ 
&$\bm{0.859\pm0.007}$ &$\bm{1.711\pm0.289}$
&$\bm{0.901\pm0.019}$ &$\bm{4.531\pm1.382}$\\
\bottomrule
\end{tabular*}}
\label{Comparison}
\end{table*}

\begin{table*}
\centering
\renewcommand\arraystretch{1.1}
\caption{Comparison results of reg-model on the three datasets. The other four state-of-the-art registration methods are included in the first four rows. The best performance is marked in bold}
\scalebox{1.0}{
\begin{tabular*}{0.9\linewidth}{lcccccc}
\toprule
\multirow{1}*{\centering Method} & \multicolumn{2}{c}{\centering OASIS} & \multicolumn{2}{c}{\centering CANDIShare} & \multicolumn{2}{c}{MM-WHS 2017}\\
\cmidrule(lr){2-3} \cmidrule(lr){4-5}\cmidrule(lr){6-7
}
& Dice $\uparrow$  & HD $\downarrow$  & Dice $\uparrow$  & HD $\downarrow$ & Dice $\uparrow$  & HD $\downarrow$\\
\midrule
Elastix &$0.750\pm0.019$ &$1.852\pm0.354$ & $0.785\pm0.016$ &$2.232\pm0.348$ &$0.746\pm0.102$ &$9.639\pm4.901$\\
VoxelMorph &$0.787\pm0.019$ &$1.625\pm0.271$ &$0.802\pm0.018$ &$2.229\pm0.430$ &$0.811\pm0.046$ &$7.850\pm2.584$\\
10*VTN &$0.798\pm0.018$ &$1.497\pm0.278$ &$0.805\pm0.015$ &$2.092\pm0.319$ &$0.874\pm0.023$ &$6.640\pm1.872$\\
PCNet &$0.808\pm0.014$ &$1.453\pm0.250$ &$0.812\pm0.013$ &$2.110\pm0.347$ &$0.850\pm0.040$ &$6.794\pm2.482$\\
\midrule
Brainstorm &$0.782\pm0.024$ &$1.674\pm0.331$ &$0.800\pm0.026$ &$2.325\pm0.599$ &$0.812\pm0.044$ &$7.268\pm2.142$\\
DeepAtlas &$0.800\pm0.019$ &$1.638\pm0.303$ &$0.818\pm0.014$ &$2.096\pm0.328$ &$0.843\pm0.041$ &$6.654\pm2.190$\\
PC-Reg-RT &$0.731\pm0.033$ &$2.072\pm0.376$ &$0.819\pm0.013$ &$2.132\pm0.355$ &$0.832\pm0.042$ &$6.587\pm1.980$\\
BRBS &$0.796\pm0.011$ &$1.579\pm0.211$ &$0.824\pm0.009$ &$2.044\pm0.329$ &$0.879\pm0.023$ &$5.099\pm1.346$\\
StyleSeg &$0.841\pm0.016$ &$1.520\pm0.272$ &$0.831\pm0.013$ &$1.981\pm0.279$ &$0.880\pm0.025$ &$5.102\pm1.455$\\
StyleSeg~V2 &$\bm{0.847\pm0.004}$ &$\bm{1.330\pm0.239}$ 
&$\bm{0.844\pm0.005}$ &$\bm{1.928\pm0.297}$
&$\bm{0.894\pm0.020}$ &$\bm{4.647\pm1.453}$\\
\bottomrule
\end{tabular*}}
\label{Comparison_reg}
\end{table*}

\begin{figure*}[!t]
\centerline{\includegraphics[width=0.95\textwidth]{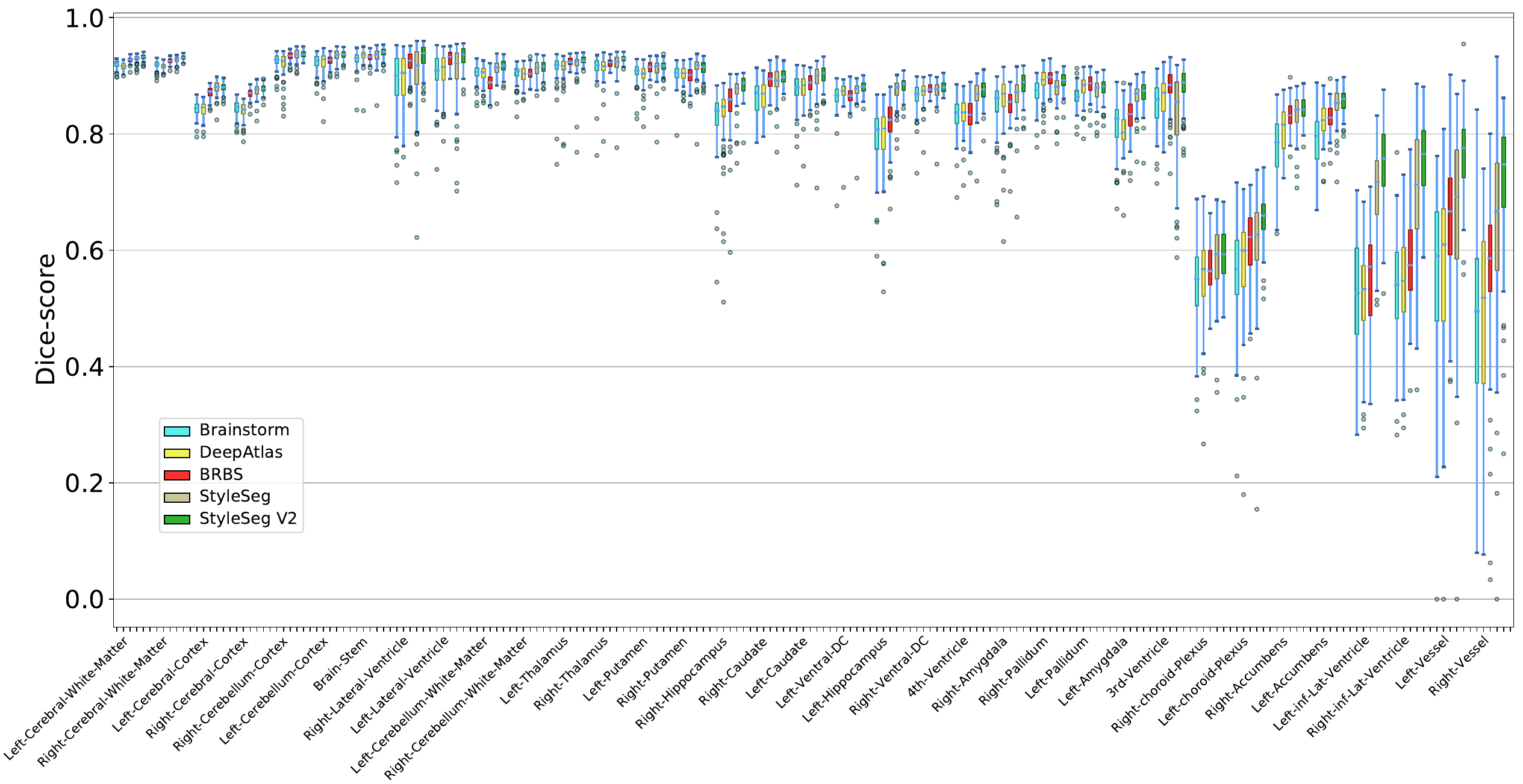}}
\caption{Boxplots of Seg-models' Dice scores of 35 cortical and subcortical brain structures on OASIS dataset. Five dual-model learning methods in Table~\ref{Comparison}, i.e., Brainstorm, Deepatlas, BRBS, StyleSeg and our proposed StyleSeg~v2 are compared. The names of such brain structures are ranked by the average numbers of the region voxels in decreasing order.}
\label{boxplot}
\end{figure*}

Therefore, the whole loss function can be formulated as:
\begin{equation}\label{Registration loss}
    \mathcal{L}_{Reg}^{i}=\left\{
	\begin{aligned}
	&\mathcal{L}_{IC} + \mathcal{L}_{Smo}, \quad \text{if} \enspace i=0\\
	&\mathcal{L}_{IC} + \mathcal{L}_{weak} + \mathcal{L}_{Smo}, \quad \text{if} \enspace i\geq 1.\\
	\end{aligned}
	\right
	.
\end{equation}
where $i$ indexes the iterations and $0$ means the initial phase. In our experiment, we utilize 3 interations of training to get the final model.

\begin{figure*}[!t]
\centerline{\includegraphics[width=0.95\textwidth]{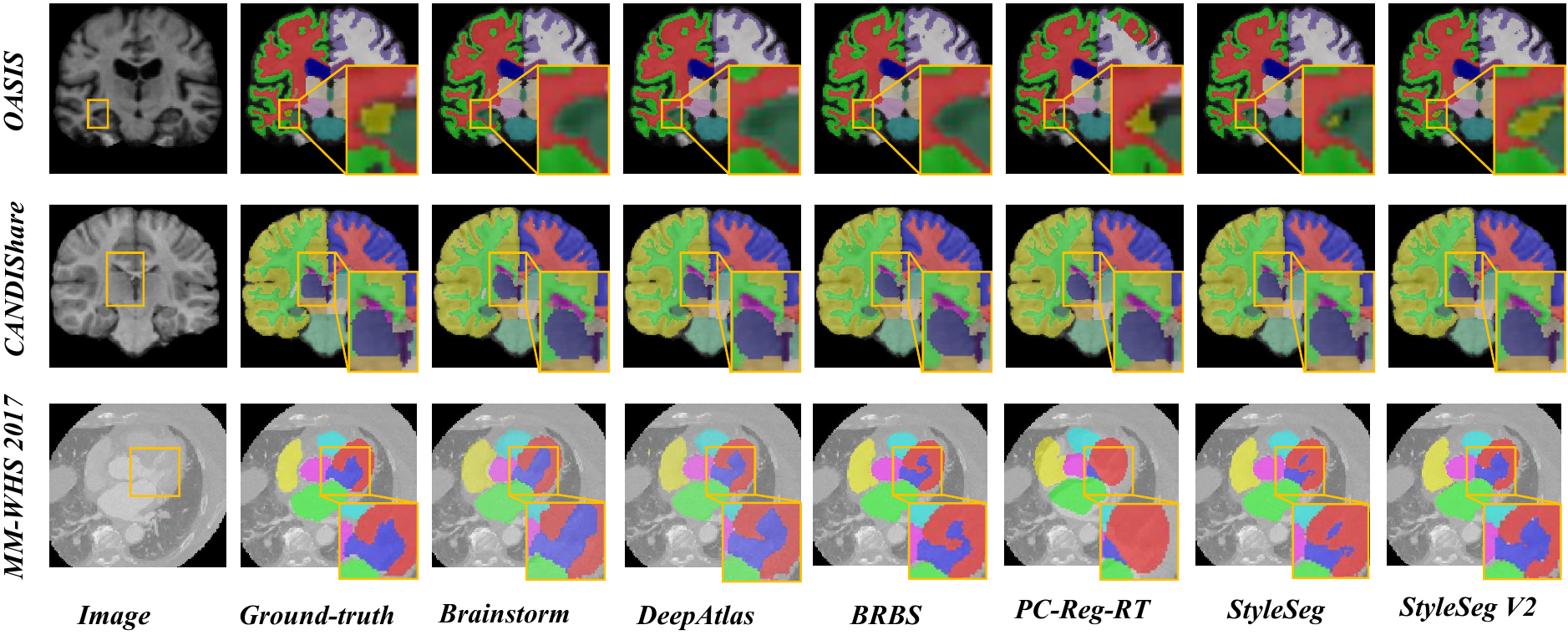}}
\caption{The visualization results of different dual-model based one-shot segmentation methods (the seg-model part) on the three datasets.}
\label{VIS}
\end{figure*}

\subsubsection{Segmentation constraints} We use two losses  $\mathcal{L}_{d}$ and $\mathcal{L}_{cgd}$ as explained in Sec.~\ref{l_{d}} and Sec.~\ref{Lcgd}, respectively, for supervising the seg-model. The whole loss function can be calculated as:
\begin{equation}\label{Segmentation loss}
    \mathcal{L}_{Seg} = \mathcal{L}_{d} + \lambda\mathcal{L}_{cgd},
\end{equation}
where $\lambda$ is set to 0.5 in our experiment.

For the reg-model in each iteration, the learning rate is set to $1 \times 10^{-4}$, the batch size is 1, and the number of training steps is 20,000. For the seg-model in each iteration, the learning rate, batch size and the number of training steps are set to $1 \times 10^{-3}$, 1, and 10,000, respectly. During training, we perform random spatial transformations including affine and B-spline transformations on each training image to enhance the robustness. We implement our method based on Tensorflow \cite{tf} and use the Adam optimizer to train the network. The number of iterations is set to 3. All training and testing are performed using a GPU resource of NVIDIA RTX A6000 with 48-GB memory.

\section{DATASET AND EVALUATION METRICS}
\subsection{Dataset}
We use two brain MRI datasets OASIS \cite{marcus2007open} and CANDIShare \cite{kennedy2012candishare} and one heart CT dataset MH-WHS 2017 \cite{zhuang2016multi} for evaluation and comparison in this work.

\subsubsection{OASIS} The dataset contains 414 scans of T1 brain MRI with an image size of 256×256×256 and a voxel spacing of 1×1×1 mm. The ground-truth masks for 35 brain tissues are obtained by FreeSufer \cite{fischl2012freesurfer} and SAMSEG \cite{puonti2016fast}. All brain MRI images were pre-processed to remove the skull, corrected the bias field and rigidly preregistered.

\subsubsection{CANDIShare} The dataset contains 103 scans of T1 brain MRI with an image size ranging from 256×256×128 to 256×256×158, and the voxel spacing is around 1×1×1.5 mm. The dataset also provides labeled data of multiple brain structures. We selected 28 brain tissues for experiments referred as \cite{wang2020lt} for a fair comparison. We performed a same pre-processing operation as OASIS on this dataset.

\subsubsection{MM-WHS 2017} contains 20 labeled heart CT images with the GT mask of 7 big cardiac structures, and 40 unlabeled images. The images are also rigid-aligned with each other using the similar data pre-processing.

For the brain MRI datasets, we randomly divide the data in each dataset into training and test sets, obtaining 331 training and 83 test images in OASIS, and 83 training and 20 test images in CANDIShare. We select an image from the training set as atlas in both OASIS and CANDIShare by following \cite{wang2020lt,puonti2016fast}, and the rest of the training set is viewed as unlabeled images. For the CT heart images, the 40 unlabeled images are used as the training set, and a labeled image is randomly selected as atlas and the rest 19 are viewed as the test set.

\subsection{Evaluation metrics}
We use Dice score and Hausdorff Distance (HD) as evaluation metrics. The calculation of Dice is formulated in Eq.~(\ref{Dice}). The Hausdorff Distance measures the maximum mismatch between surfaces of two volumes, which is formulated as:
\begin{equation}\label{HD}
    HD(A, B) = \mathop{\max}_{a \in A} \left\{ \mathop{\min}_{b \in B}\left\{ || a-b || \right\} \right\}.
\end{equation}
where A and B indicate all voxels on the surfaces of two volumes respectively.

We also perform paired sample T test \cite{biometrika1908probable} using Excel software for every comparison in the following. If the p-values are less than or equal to 0.001, the improvements are considered to be statistically significant.

\section{EXPERIMENTAL RESULTS}
\subsection{Comparison with the state-of-the-arts}
\subsubsection{Performance of Seg-model}
We first compare the seg-model's performance of StyleSeg~V2 with 5 state-of-the-art dual-model learning methods, i.e., Brainstorm \cite{zhao2019data}, DeepAtlas \cite{xu2019deepatlas}, PC-Reg-RT \cite{he2021few}, BRBS \cite{he2022learning} and StyleSeg \cite{lv2023robust}. We directly utilize their released source code for comparison. Table~\ref{Comparison} shows the average Dice score and Hausdorff distance of seg-model on OASIS, CANDIShare and MM-WHS 2017 datasets. As can be seen, StyleSeg~V2 outperforms all comparison methods by different margins. Benefiting from the new ability of registration error perception, StyleSeg~V2 exceeds its previous version StyleSeg by increasing Dice by 2.0\%, 2.4\%, and 1.7\% on the three datasets, respectively. Besides, StyleSeg~V2 also achieves a superior performance over BRBS by increasing Dice by 4.0\%, 0.9\%, and 1.0\% on the three datasets, respectively. All the improvements are verified to be statistically significant ($p<0.001$). Fig.~\ref{boxplot} shows boxplots of the Dice scores in 35 cortical regions on OASIS, obtained by top five different methods in Table~\ref{Comparison}. As can be seen, a superior performance is consistently achieved by our method in every brain region. Moreover, the Dice scores obtained by our method are more stable as the boxes are shorter in length, which verifies the robustness.

We also visualize the segmentation results of all comparison methods on the three datasets in Fig.~\ref{VIS}. From the regions highlighted in the boxes, we can observe that the predicted masks of StyleSeg~V2 are most consistent with GT across different modalities and organs.

\subsubsection{Performance of Reg-model}
The seg-model is good at specific regions, while the reg-model also matters when facing the registration-based tasks like intraoperative navigation and arbitrary tissue segmentation. Therefore, we also evaluate the reg-model of StyleSeg~V2, and include 9 state-of-the-arts for comparison. The evaluation is performed by viewing these comparison registration models as atlas-based segmentation, and also using Dice and HD of labeled regions as evaluation metrics. 
Table~\ref{Comparison_reg} lists the comparison results, where some state-of-the-art architectures for merely brain registration, i.e., Elastix~\cite{klein2009elastix}, VoxelMorph~\cite{balakrishnan2019voxelmorph}, 10*VTN~\cite{Zhao_2019_ICCV} and PCNet~\cite{lv2022joint}, are also included.

As can be seen, the reg-model of StyleSeg V2 outperforms that of the original StyleSeg by increasing Dice by 0.7\%, 1.6\% and 1.6\% respectively and greatly exceeds other registration and dual-model learning methods. 
Besides, we observe that the segmentation performances of other dual-model learning methods are almost on par with, and even inferior to the pure registration methods (the first four rows). 
On contrary, StyleSeg~V2 can achieve `win-win' results of both seg-model and reg-model, showing greater potentials for both segmentation of specific brain tissues and other registration-based applications.

\subsection{Ablation studies}
We conduct ablation studies on OASIS to verify the effectiveness of our proposed two modules enabled by the perceived registration errors, i.e., WIST and $\mathcal{L}_{cgd}$, and also to explore the impact of division quantity $N$ on WIST.

\begin{figure}[t]
\centerline{\includegraphics[width=0.9\columnwidth]{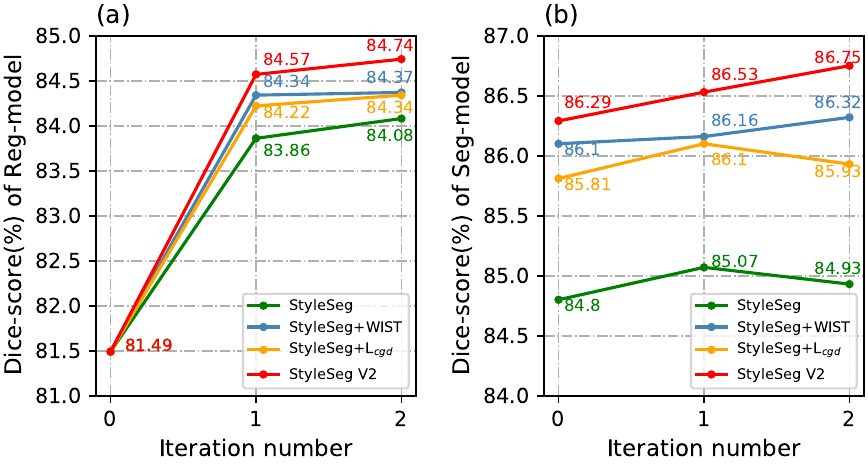}}
    \caption{Line graphs of the average Dice of different variants at each iteration. (a) the line graphs of reg-model, and (b) the line graphs of seg-model.}
	\label{polyline}
\end{figure}

\begin{figure}[t]
\centerline{\includegraphics[width=1.0\columnwidth]{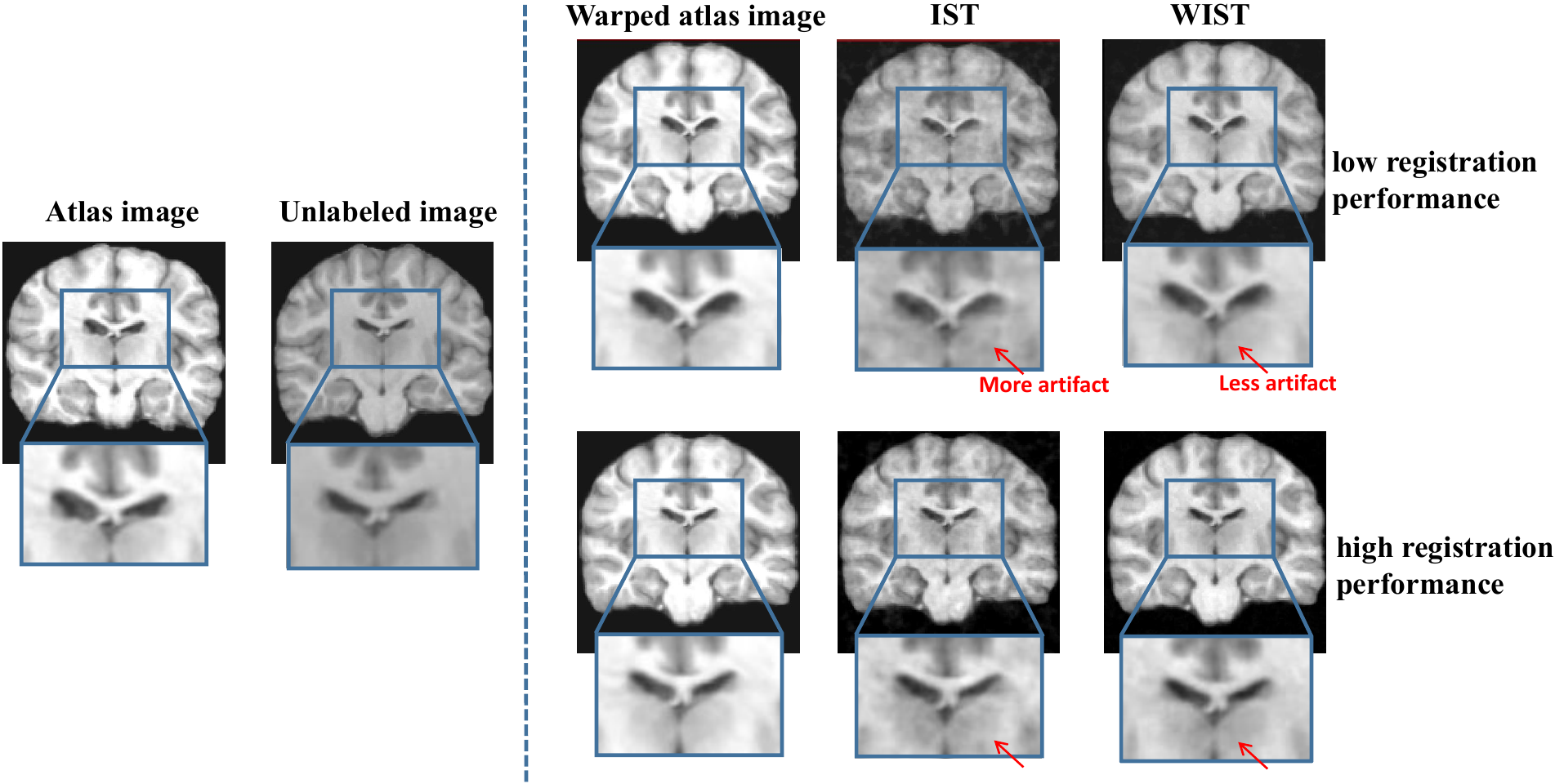}}
    \caption{Comparison result of IST and WIST. By use of WIST, fewer artifacts are produced as highlighted in the blue box.}
	\label{IST_WIST}
\end{figure}

\subsubsection{Effectiveness of WIST and $\mathcal{L}_{cgd}$} 
To verify the effectiveness of WIST and $\mathcal{L}_{cgd}$, We train three variants of StyleSeg~V2, denoted as ``StyleSeg'', ``StyleSeg+WIST'', and ``StyleSeg+$\mathcal{L}_{cgd}$'', by disabling WIST and/or $\mathcal{L}_{cgd}$. The StyleSeg~V2 itself is actually ``StyleSeg+WIST\&$\mathcal{L}_{cgd}$''. 
If disabling WIST, we use the original IST for style transformation. If disabling $\mathcal{L}_{cgd}$, we discard the unlabeled images in the segmentation training.
We show the performance of reg-model and seg-model at each iteration, and plot the line graphs in Fig.~\ref{polyline}. 

From Fig.~\ref{polyline}, we have four observations:

(i) Comparing the results of ``StyleSeg+WIST'' with ``StyleSeg'' in Fig.~\ref{polyline}(b), we can observe that after 3 iterations, the seg-model with WIST achieves an improvement of average Dice by 1.6\% over the original StyleSeg. Fig.~\ref{IST_WIST} shows exemplar results of IST and WIST with $\beta$ set as the maximum. We can see that when the registration accuracy is relatively unsatisfactory in the initial training phase, the subtle structures generated via IST is damaged with artifacts, which may degrade the subsequent segmentation learning. In comparison, WIST induces fewer artifacts without sacrificing the effect of style transformation under both low and high registration performance. These results show our WIST can produce style-diverse data with less artifacts and effectively enhance the segmentation and registration performance of our iterative dual-model.

(ii) Comparing the results of 
``StyleSeg+$\mathcal{L}_{cgd}$'' with ``StyleSeg'' in Fig.~\ref{polyline}(b), we can observe the seg-model with $\mathcal{L}_{cgd}$ finally achieves an improvement of average Dice by 1.2\%. This verifies the non-ignorable value of unlabeled images in the segmentation training, which, however, has to be abandoned in the original StyleSeg due to the misalignment caused by the reg-model.

(iii) Comparing the results of ``StyleSeg+WIST'' with ``StyleSeg+$\mathcal{L}_{cgd}$'' in Fig.~\ref{polyline}(b), we can observe that applying WIST only boosts the segmentation performance more significantly than applying $\mathcal{L}_{cgd}$ only. 
This indicates that the alignment between training image and mask is more important for the dual-model iterative learning.
Therefore, addressing the diversity and fidelity of the warped atlas image may be the promising direction to enhance the performance of dual-model based one-shot brain segmentation.

(iv) Comparing the Fig.~\ref{polyline}(a) with (b), we can observe that our proposed WIST and $\mathcal{L}_{cgd}$ also help to improve the performance of reg-model, which in turn boosts the seg-model during iteration.

\subsubsection{The impact of division quantity $N$ on WIST} In WIST, we divide the confidence map $C$ into $N$ parts, but the determination of $N$ is a trade-off: when $N$ is small, the style contrast of different regions becomes intense, yielding visually unreal boundaries, and when $N$ is large, the interval $[\frac{n}{N}, \frac{n+1}{N})$ becomes narrow, and the random selection of the corresponding $\beta_{n}$ becomes more certain, which lowers the diversity of style-transformed
atlas image.

\begin{figure}[t]
\centerline{\includegraphics[width=1.0\columnwidth]{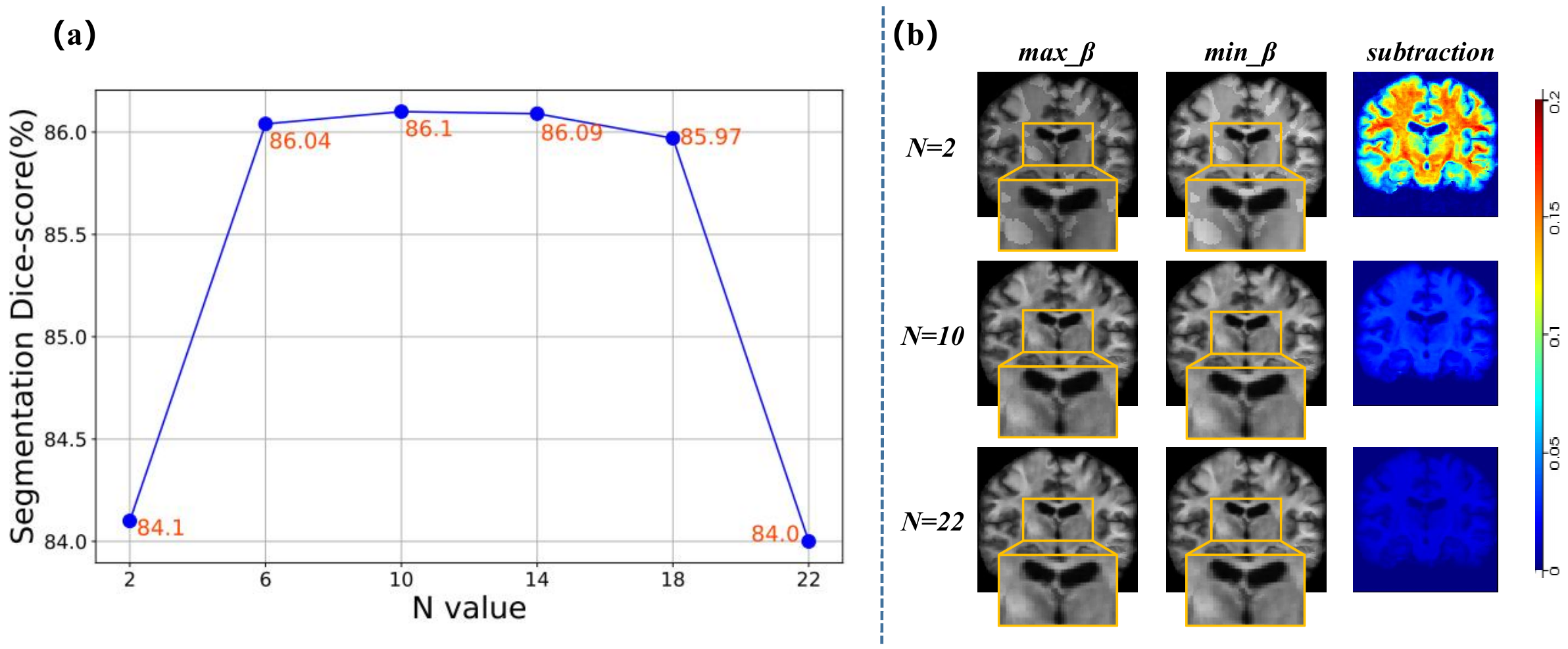}}
    \caption{The impact of different sets of $N$ in WIST. (a) the line graph of seg-model's Dice at different $N$, and (b) subtraction images reflecting the diversity of generated style-transformed atlas image (more brighter, more diverse).}
	\label{wist_num}
\end{figure}

In view of this, we train 6 variants of StyleSeg~V2 with $N$ in WIST set to that uniformly-space sampled from $2$ to $22$. Fig.~\ref{wist_num}(a) shows the segmentation performance of seg-model at the first iteration. As can be seen, the seg-model will degrade when employing both too small and too large $N$ value. To explain this phenomenon, we set $\beta_{n}$ to the rightmost and leftmost values of each interval to generate two style-transformed atlas images, and calculate their subtraction to reflect the style diversity under different selections of $N$. Fig.~\ref{wist_num}(b) shows the style-transformed atlas images using the rightmost and leftmost values as $\beta_{n}$, and their subtraction images. As can be seen, using smaller $N$ provides a greater space where the style varies (brighter subtraction image), but also creates more salient unreal boundaries, which harms the image structures. On the contrary, using larger $N$ produces more realistic image but narrows the style varying space (darker subtraction image). To trade-off the diversity and fidelity, we set $N$ to 10 in StyleSeg~V2 by default.

\subsection{Qualitative verification of optimization-free error perception}
\begin{figure}[t]
\centerline{\includegraphics[width=0.95\columnwidth]{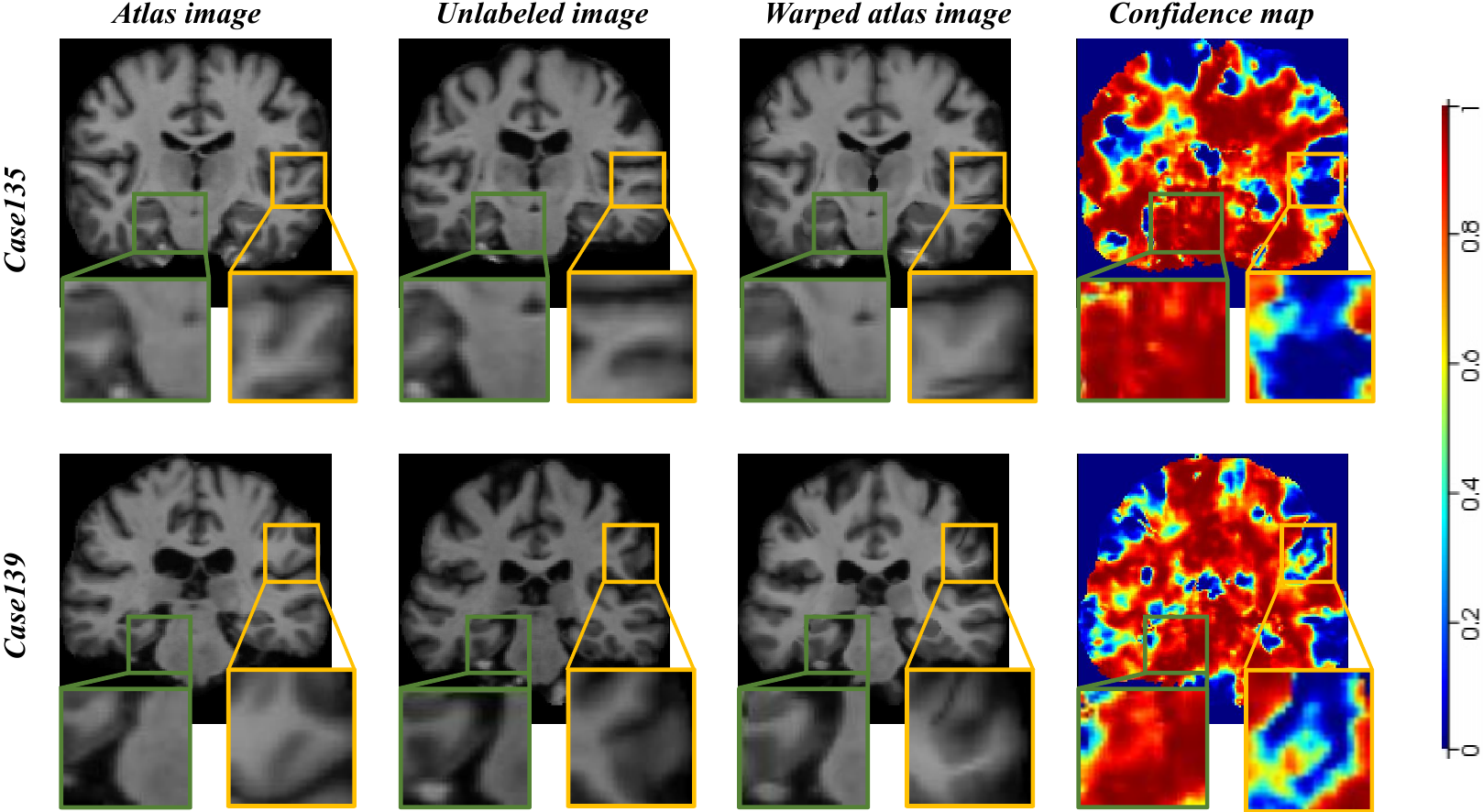}}
    \caption{Visualization of the perceived confidence map using our proposed optimization-free manner, where the red means high confidence while the blue means low confidence. The green and yellow boxes zoom in on the regions with high and low confidence values, respectively.} 
	\label{error}
\end{figure}

To visually verify the rationality of our optimization-free error perception via the idea of mirroring the data, we randomly select two cases from OASIS test set, and calculate the confidence map using our proposed optimization-free registration error perception as mentioned in Sec.~\ref{mirror_error}. Fig.~\ref{error} shows the results. As can be seen, the high confident regions (i.e., the red in confidence map) indicate better registration performance, which can be proved by comparing the areas highlighted in the green box. On the contrary, the low confident regions (i.e., the blue in confidence map) have a relatively unsatisfactory registration performance, which can be proved by comparing the areas highlighted in the yellow box. These results can verify that the mirror error can reflect the registration performance, and without requiring any learnable parameter or optimization step, it can be utilized to guide a better style-transformation in WIST and more reliable supervisions using the unlabeled images' pseudo masks, and ultimately alleviates the degradation of segmentation performance caused by registration error.

\section{CONCLUSION}
In this paper, we propose StyleSeg~V2 for robust one-shot brain segmentation, which distinguishes itself from the original StyleSeg by a new ability of perceiving registration errors without using any learnable extra model. On the top of registration error perception, we reinvent the image-aligned style transformation (IST) as weighted IST (WIST), which can improve both the diversity and fidelity of the style-transformed warped copies of atlas image. Moreover, we introduce a confidence guided Dice-loss ($\mathcal{L}_{cgd}$) to make the segmentation model directly benefit from those correctly-aligned regions of unlabeled images. Thanks to the better style-transformation and more reliable pseudo-supervisions, our proposed StyleSeg V2 can play the full effect of both labeled altas and unlabeled images, and absorb their carried knowledge more effectively. We conduct extensive and comprehensive experiments including two public brain datasets, i.e., OASIS and CANDIShare, and one heart dataset, i.e., MM-WHS 2017. Comparison results on such three datasets demonstrate that using only a single labeled image for training, StyleSeg~V2 outperforms its previous version StyleSeg and other state-of-the-art methods for both registration and segmentation performance by considerable margins. The ablation studies verify the effectiveness of our proposed two components, i.e., WIST and confidence guided Dice-loss, and also make a discussion over the determination of division quantity N. The qualitative verification experiment of optimization-free error perception visually reveals a favorable correspondence between mirror error and registration error. In the future work, we will further verify StyleSeg~V2 in preoperative-postoperative registration where perceiving registration errors is of great importance. 

\bibliographystyle{IEEEtran}
\bibliography{IEEEabrv,ref}

\begin{thebibliography}{10}
\providecommand{\url}[1]{#1}
\csname url@samestyle\endcsname
\providecommand{\newblock}{\relax}
\providecommand{\bibinfo}[2]{#2}
\providecommand{\BIBentrySTDinterwordspacing}{\spaceskip=0pt\relax}
\providecommand{\BIBentryALTinterwordstretchfactor}{4}
\providecommand{\BIBentryALTinterwordspacing}{\spaceskip=\fontdimen2\font plus
\BIBentryALTinterwordstretchfactor\fontdimen3\font minus \fontdimen4\font\relax}
\providecommand{\BIBforeignlanguage}[2]{{%
\expandafter\ifx\csname l@#1\endcsname\relax
\typeout{** WARNING: IEEEtran.bst: No hyphenation pattern has been}%
\typeout{** loaded for the language `#1'. Using the pattern for}%
\typeout{** the default language instead.}%
\else
\language=\csname l@#1\endcsname
\fi
#2}}
\providecommand{\BIBdecl}{\relax}
\BIBdecl

\bibitem{geuze2005mr}
E.~Geuze, E.~Vermetten, and J.~D. Bremner, ``Mr-based in vivo hippocampal volumetrics: 2. findings in neuropsychiatric disorders,'' \emph{Molecular psychiatry}, vol.~10, no.~2, pp. 160--184, 2005.

\bibitem{lorenzo2002atlas}
M.~Lorenzo-Vald{\'e}s, G.~I. Sanchez-Ortiz, R.~Mohiaddin, and D.~Rueckert, ``Atlas-based segmentation and tracking of 3d cardiac mr images using non-rigid registration,'' in \emph{Medical Image Computing and Computer-Assisted Intervention—MICCAI 2002: 5th International Conference Tokyo, Japan, September 25--28, 2002 Proceedings, Part I 5}.\hskip 1em plus 0.5em minus 0.4em\relax Springer, 2002, pp. 642--650.

\bibitem{lotjonen2010fast}
J.~M. L{\"o}tj{\"o}nen, R.~Wolz, J.~R. Koikkalainen, L.~Thurfjell, G.~Waldemar, H.~Soininen, D.~Rueckert, A.~D.~N. Initiative \emph{et~al.}, ``Fast and robust multi-atlas segmentation of brain magnetic resonance images,'' \emph{Neuroimage}, vol.~49, no.~3, pp. 2352--2365, 2010.

\bibitem{he2020deep}
Y.~He, T.~Li, G.~Yang, Y.~Kong, Y.~Chen, H.~Shu, J.-L. Coatrieux, J.-L. Dillenseger, and S.~Li, ``Deep complementary joint model for complex scene registration and few-shot segmentation on medical images,'' in \emph{Computer Vision--ECCV 2020: 16th European Conference, Glasgow, UK, August 23--28, 2020, Proceedings, Part XVIII 16}.\hskip 1em plus 0.5em minus 0.4em\relax Springer, 2020, pp. 770--786.

\bibitem{beljaards2020cross}
L.~Beljaards, M.~S. Elmahdy, F.~Verbeek, and M.~Staring, ``A cross-stitch architecture for joint registration and segmentation in adaptive radiotherapy,'' in \emph{Medical Imaging with Deep Learning}.\hskip 1em plus 0.5em minus 0.4em\relax PMLR, 2020, pp. 62--74.

\bibitem{zhao2019data}
A.~Zhao, G.~Balakrishnan, F.~Durand, J.~V. Guttag, and A.~V. Dalca, ``Data augmentation using learned transformations for one-shot medical image segmentation,'' in \emph{Proceedings of the IEEE/CVF conference on computer vision and pattern recognition}, 2019, pp. 8543--8553.

\bibitem{he2022learning}
Y.~He, R.~Ge, X.~Qi, Y.~Chen, J.~Wu, J.-L. Coatrieux, G.~Yang, and S.~Li, ``Learning better registration to learn better few-shot medical image segmentation: Authenticity, diversity, and robustness,'' \emph{IEEE Transactions on Neural Networks and Learning Systems}, 2022.

\bibitem{lv2023robust}
J.~Lv, X.~Zeng, S.~Wang, R.~Duan, Z.~Wang, and Q.~Li, ``Robust one-shot segmentation of brain tissues via image-aligned style transformation,'' in \emph{Proceedings of the AAAI Conference on Artificial Intelligence}, vol.~37, no.~2, 2023, pp. 1861--1869.

\bibitem{collins1995automatic}
D.~L. Collins, C.~J. Holmes, T.~M. Peters, and A.~C. Evans, ``Automatic 3-d model-based neuroanatomical segmentation,'' \emph{Human brain mapping}, vol.~3, no.~3, pp. 190--208, 1995.

\bibitem{suh2013automatic}
J.~W. Suh, M.~Schaap, A.~Lee, N.~Do, A.~Ahiekpor-Dravi, Y.~Bai, G.~Choi, and R.~Moreau-Gobard, ``Automatic multi-atlas segmentation using dual registrations,'' in \emph{2013 IEEE 10th International Symposium on Biomedical Imaging}.\hskip 1em plus 0.5em minus 0.4em\relax IEEE, 2013, pp. 1284--1287.

\bibitem{avants2008symmetric}
B.~B. Avants, C.~L. Epstein, M.~Grossman, and J.~C. Gee, ``Symmetric diffeomorphic image registration with cross-correlation: evaluating automated labeling of elderly and neurodegenerative brain,'' \emph{Medical image analysis}, vol.~12, no.~1, pp. 26--41, 2008.

\bibitem{klein2009elastix}
S.~Klein, M.~Staring, K.~Murphy, M.~A. Viergever, and J.~P. Pluim, ``Elastix: a toolbox for intensity-based medical image registration,'' \emph{IEEE transactions on medical imaging}, vol.~29, no.~1, pp. 196--205, 2009.

\bibitem{wang2020lt}
S.~Wang, S.~Cao, D.~Wei, R.~Wang, K.~Ma, L.~Wang, D.~Meng, and Y.~Zheng, ``Lt-net: Label transfer by learning reversible voxel-wise correspondence for one-shot medical image segmentation,'' in \emph{Proceedings of the IEEE/CVF Conference on Computer Vision and Pattern Recognition}, 2020, pp. 9162--9171.

\bibitem{ding2022aladdin}
Z.~Ding and M.~Niethammer, ``Aladdin: Joint atlas building and diffeomorphic registration learning with pairwise alignment,'' in \emph{Proceedings of the IEEE/CVF conference on computer vision and pattern recognition}, 2022, pp. 20\,784--20\,793.

\bibitem{dual-MICCAI}
X.~Hu, M.~Kang, W.~Huang, M.~R. Scott, R.~Wiest, and M.~Reyes, ``Dual-stream pyramid registration network,'' in \emph{International Conference on Medical Image Computing and Computer-Assisted Intervention}.\hskip 1em plus 0.5em minus 0.4em\relax Springer, 2019, pp. 382--390.

\bibitem{mok2020large}
T.~C. Mok and A.~C. Chung, ``Large deformation diffeomorphic image registration with laplacian pyramid networks,'' in \emph{Medical Image Computing and Computer Assisted Intervention--MICCAI 2020: 23rd International Conference, Lima, Peru, October 4--8, 2020, Proceedings, Part III 23}.\hskip 1em plus 0.5em minus 0.4em\relax Springer, 2020, pp. 211--221.

\bibitem{zhao2019unsupervised}
S.~Zhao, T.~Lau, J.~Luo, I.~Eric, C.~Chang, and Y.~Xu, ``Unsupervised 3d end-to-end medical image registration with volume tweening network,'' \emph{IEEE journal of biomedical and health informatics}, vol.~24, no.~5, pp. 1394--1404, 2019.

\bibitem{Zhao_2019_ICCV}
S.~Zhao, Y.~Dong, E.~I.-C. Chang, and Y.~Xu, ``Recursive cascaded networks for unsupervised medical image registration,'' in \emph{Proceedings of the IEEE/CVF International Conference on Computer Vision (ICCV)}, October 2019.

\bibitem{hu2022recursive}
B.~Hu, S.~Zhou, Z.~Xiong, and F.~Wu, ``Recursive decomposition network for deformable image registration,'' \emph{IEEE Journal of Biomedical and Health Informatics}, vol.~26, no.~10, pp. 5130--5141, 2022.

\bibitem{lv2022joint}
J.~Lv, Z.~Wang, H.~Shi, H.~Zhang, S.~Wang, Y.~Wang, and Q.~Li, ``Joint progressive and coarse-to-fine registration of brain mri via deformation field integration and non-rigid feature fusion,'' \emph{IEEE Transactions on Medical Imaging}, vol.~41, no.~10, pp. 2788--2802, 2022.

\bibitem{xu2019deepatlas}
Z.~Xu and M.~Niethammer, ``Deepatlas: Joint semi-supervised learning of image registration and segmentation,'' in \emph{Medical Image Computing and Computer Assisted Intervention--MICCAI 2019: 22nd International Conference, Shenzhen, China, October 13--17, 2019, Proceedings, Part II 22}.\hskip 1em plus 0.5em minus 0.4em\relax Springer, 2019, pp. 420--429.

\bibitem{eppenhof2017supervised}
K.~A. Eppenhof and J.~P. Pluim, ``Supervised local error estimation for nonlinear image registration using convolutional neural networks,'' in \emph{Medical Imaging 2017: Image Processing}, vol. 10133.\hskip 1em plus 0.5em minus 0.4em\relax SPIE, 2017, pp. 526--531.

\bibitem{eppenhof2018error}
{Eppenhof, Koen AJ and Pluim, Josien PW}, ``Error estimation of deformable image registration of pulmonary ct scans using convolutional neural networks,'' \emph{Journal of medical imaging}, vol.~5, no.~2, pp. 024\,003--024\,003, 2018.

\bibitem{sokooti2021hierarchical}
H.~Sokooti, S.~Yousefi, M.~S. Elmahdy, B.~P. Lelieveldt, and M.~Staring, ``Hierarchical prediction of registration misalignment using a convolutional lstm: Application to chest ct scans,'' \emph{IEEE Access}, vol.~9, pp. 62\,008--62\,020, 2021.

\bibitem{bierbrier2023towards}
J.~Bierbrier, M.~Eskandari, D.~A. Di~Giovanni, and D.~L. Collins, ``Towards estimating mri-ultrasound registration error in image-guided neurosurgery,'' \emph{IEEE Transactions on Ultrasonics, Ferroelectrics, and Frequency Control}, 2023.

\bibitem{christensen2001consistent}
G.~E. Christensen and H.~J. Johnson, ``Consistent image registration,'' \emph{IEEE transactions on medical imaging}, vol.~20, no.~7, pp. 568--582, 2001.

\bibitem{meister2018unflow}
S.~Meister, J.~Hur, and S.~Roth, ``Unflow: Unsupervised learning of optical flow with a bidirectional census loss,'' in \emph{Proceedings of the AAAI conference on artificial intelligence}, vol.~32, no.~1, 2018.

\bibitem{mok2022unsupervised}
T.~C. Mok and A.~C. Chung, ``Unsupervised deformable image registration with absent correspondences in pre-operative and post-recurrence brain tumor mri scans,'' in \emph{International Conference on Medical Image Computing and Computer-Assisted Intervention}.\hskip 1em plus 0.5em minus 0.4em\relax Springer, 2022, pp. 25--35.

\bibitem{saygili2015confidence}
G.~Saygili, M.~Staring, and E.~A. Hendriks, ``Confidence estimation for medical image registration based on stereo confidences,'' \emph{IEEE transactions on medical imaging}, vol.~35, no.~2, pp. 539--549, 2015.

\bibitem{tf}
M.~Abadi, P.~Barham, J.~Chen, Z.~Chen, A.~Davis, J.~Dean, M.~Devin, S.~Ghemawat, G.~Irving, M.~Isard \emph{et~al.}, ``$\{$TensorFlow$\}$: a system for $\{$Large-Scale$\}$ machine learning,'' in \emph{12th USENIX symposium on operating systems design and implementation (OSDI 16)}, 2016, pp. 265--283.

\bibitem{marcus2007open}
D.~S. Marcus, T.~H. Wang, J.~Parker, J.~G. Csernansky, J.~C. Morris, and R.~L. Buckner, ``Open access series of imaging studies (oasis): cross-sectional mri data in young, middle aged, nondemented, and demented older adults,'' \emph{Journal of cognitive neuroscience}, vol.~19, no.~9, pp. 1498--1507, 2007.

\bibitem{kennedy2012candishare}
D.~N. Kennedy, C.~Haselgrove, S.~M. Hodge, P.~S. Rane, N.~Makris, and J.~A. Frazier, ``Candishare: a resource for pediatric neuroimaging data,'' \emph{Neuroinformatics}, vol.~10, pp. 319--322, 2012.

\bibitem{zhuang2016multi}
X.~Zhuang and J.~Shen, ``Multi-scale patch and multi-modality atlases for whole heart segmentation of mri,'' \emph{Medical image analysis}, vol.~31, pp. 77--87, 2016.

\bibitem{fischl2012freesurfer}
B.~Fischl, ``Freesurfer,'' \emph{Neuroimage}, vol.~62, no.~2, pp. 774--781, 2012.

\bibitem{puonti2016fast}
O.~Puonti, J.~E. Iglesias, and K.~Van~Leemput, ``Fast and sequence-adaptive whole-brain segmentation using parametric bayesian modeling,'' \emph{NeuroImage}, vol. 143, pp. 235--249, 2016.

\bibitem{biometrika1908probable}
S.~Biometrika, ``The probable error of a mean,'' \emph{Biometrika}, vol.~6, no.~1, pp. 1--25, 1908.

\bibitem{he2021few}
Y.~He, T.~Li, R.~Ge, J.~Yang, Y.~Kong, J.~Zhu, H.~Shu, G.~Yang, and S.~Li, ``Few-shot learning for deformable medical image registration with perception-correspondence decoupling and reverse teaching,'' \emph{IEEE Journal of Biomedical and Health Informatics}, vol.~26, no.~3, pp. 1177--1187, 2021.

\bibitem{balakrishnan2019voxelmorph}
G.~Balakrishnan, A.~Zhao, M.~R. Sabuncu, J.~Guttag, and A.~V. Dalca, ``Voxelmorph: a learning framework for deformable medical image registration,'' \emph{IEEE transactions on medical imaging}, vol.~38, no.~8, pp. 1788--1800, 2019.

\end{thebibliography}
\end{document}